\documentclass{article}

\usepackage[final]{corl_2022}
\usepackage[utf8]{inputenc}

\usepackage{setspace}
\usepackage{array}
\usepackage{amsmath}
\usepackage{amssymb}
\usepackage{physunits}
\usepackage{graphicx}
\usepackage{caption}
\usepackage{subcaption}
\usepackage{hyperref}
\usepackage{cleveref}
\usepackage{algorithm}
\usepackage{algpseudocode}
\usepackage{authblk}
\usepackage{soul}

\newcommand{\R}{\mathbb{R}}

\title{
 Utilizing Human Feedback for Primitive Optimization in Wheelchair Tennis}

\author[]{Arjun Krishna}
\author[]{Zulfiqar Zaidi}
\author[]{Letian Chen}
\author[]{Rohan Paleja}
\author[]{Esmaeil Seraj}
\author[]{Matthew Gombolay}

\affil[]{Georgia Institute of Technology \\\vspace{0.5em}
\texttt{\small \{akrishna49, zzaidi8, letian.chen, rohan.paleja, esmaeil.seraj\}@gatech.edu, matthew.gombolay@cc.gatech.edu}}

\begin{document}
\nolinenumbers
\maketitle


\begin{abstract}
Agile robotics presents a difficult challenge with robots moving at high speeds requiring precise and low-latency sensing and control. Creating agile motion that accomplishes the task at hand while being safe to execute is a key requirement for agile robots to gain human trust. This requires designing new approaches that are flexible and maintain knowledge over world constraints. In this paper, we consider the problem of building a flexible and adaptive controller for a challenging agile mobile manipulation task of hitting ground strokes on a wheelchair tennis robot. We propose and evaluate an extension to work done on learning striking behaviors using a probabilistic movement primitive~(ProMP) framework by
(1) demonstrating the safe execution of learned primitives on an agile mobile manipulator setup, and (2) proposing an online primitive refinement procedure that utilizes evaluative feedback from humans on the executed trajectories.
\end{abstract}

\keywords{Learning from demonstrations and feedback, Movement primitives, Agile Mobile manipulator} 


\section{Introduction}
\label{sec:intro}


Over the years, roboticists have sought to develop robots that can play various sports such as soccer~\cite{robogames,robocup}, sumo~\cite{robot_sumo}, and table tennis~\cite{buchler2022learning} to demonstrate and test the capabilities of their systems. Sporting applications serve as natural milestones for robotic systems to achieve human-level performance, as to be effective at a sport, the system needs to be able to reason about the state of the game, be agile in its response to changing observations, and be mindful of objects in its vicinity. Prior work has sought to learn such complex behaviors by either using reinforcement learning~\cite{jonas_rl_tabletennis_2021,gao_modelfree2020,abeyruwan2022isimreal} or utilizing data from an expert to train the robot.  

Learning for Demonstrations (LfD)~\cite{ARGALL2009469} is a framework for learning a policy from a set of demonstrations provided by an expert. Prior work has shown how movement primitives obtained from kinesthetic teaching can be used to teach robots to hit table tennis strokes~\cite{movementprimitives_mulling2010,mulling_momp2013,gonzalez2016prompstrike,gomez2020adaptation}. 
It has also been shown that kinesthetic demonstrations can be used successfully to teach robots to play various styles of strokes for table tennis~\cite{chen_strategyinference2020,pmlr-v155-chen21b}. 
More recently, the problem of improving a policy learned from expert demonstrations using reinforcement learning, inverse reinforcement learning, and self-supervised learning has also been explored~\cite{pmlr-v155-chen21b,chen2020bail,Cheng2018FastPL,tianli_goalseye2022}. However, most of these works are evaluated on either simulated environments or on robot arms mounted on a stationary platform.



The need for algorithms that support learning robot behavior for versatile robot platforms is exacerbated in larger-scale racket sports, such as tennis. Tennis is a more challenging problem for robots as it requires a responsive agile mobile base and higher racket head speeds than table tennis. In this work, we demonstrate the first attempt to extend the ProMP framework to an agile mobile manipulator to achieve successful tennis groundstrokes with a wheelchair tennis robot. We additionally, describe an approach to refine the learned primitives from demonstrations based on human feedback.

\section{Preliminaries}
\label{prereq}

In this section, we provide an overview of the Probabilistic Movement Primitive~(ProMP) and the notations we will use in the paper. Interested readers are encouraged to refer to \cite{gomez2020adaptation} for more details. 

The ProMP is a modeling technique that compactly represents a probability distribution over robot trajectories~\cite{paraschos2013probabilistic}. Let $q_t$ represent the joint states of the robot at time $t$. The ProMP defines a set of time-dependent basis functions (represented by $\Phi_t$) and a weight vector, $w$, that compactly encodes the robot trajectory, $\tau = \{ q_t \}$ by representing $q_t \sim \mathcal{N}(\Phi_t w, \Sigma_y)$, where $\Sigma_y$ models any white noise. 

Given a dataset of demonstrations, ProMP learns a distribution over weights, $P(w\mid\{\mu_w, \Sigma_w\}) = \mathcal{N}(\mu_w, \Sigma_w)$), that captures the common features across trajectories while factoring in the variance that captures the variation across demonstrations. The parameters, $\{ \mu_w, \Sigma_w, \Sigma_y \}$, of this Hierarchical Bayesian Model~(HBM) formulation can be computed from demonstrations by obtaining Maximum Likelihood Estimates (MLE) via exact methods or Expectation-Maximization~(EM) procedures. A key property of ProMP that we leverage in this work is to \emph{condition} it to pass through desired end-effector waypoints. Since the ProMP representation is in the joint space, some form of inverse kinematics~(IK) is required to perform this conditioning operation.




\section{Method}
\label{sec:method}
In \Cref{subsec:system}, we provide a brief overview of our experiment setup: a wheelchair tennis robot. In \Cref{subsec:stroke-controller}, we describe the details of the deployed stroke controller that safely executes the learned primitives. In \Cref{subsec:refine-prim}, we present our proposed method for improving movement primitives based on human feedback.


\subsection{System Overview}
\label{subsec:system}
We mounted a 7-DOF high-speed Barrett WAM arm on a motorized Top End Pro Tennis Wheelchair to build an agile mobile manipulator system~\cite{zaidi2022athletic}. The system is designed to emulate the athletic gameplay of regulation wheelchair tennis, where players need to react quickly in the order of a few hundred milliseconds. 

To sense and track the movement of the tennis ball, we make use of a decentralized array of stereo cameras that provides measurements from different perspectives. These estimates are fused by an Extended Kalman Filter~(EKF)~\cite{moore2016generalized} to output the ball's estimated state, which is propagated forward in time to predict the ball's future trajectory. 

The problem of whole-body control of mobile manipulators is challenging -- particularly in an agile robotics setting -- so we limit the scope of our study by constraining the wheelchair to move along one dimension. We choose to allow the robot to move laterally; notably, human players exhibit only lateral movements for the majority of the strokes~\cite{kovacs2009movement}. 
We model the lateral movement as a prismatic joint and obtain a kinematics model for the system (illustrated in \Cref{app:kinematic-model}). 



\subsection{Stroke Controller}
\label{subsec:stroke-controller}

We build our stroke controller upon ProMP, originally proposed for table tennis in~\cite{gonzalez2016prompstrike}. We make two key advances for our wheelchair tennis robot.

First, since the strokes executed often reach racket-head speeds  $\sim10$\mps, it is critical to ensure that the conditioned joint space configuration at the time of impact is safe and achievable without any self-collisions. Thus, we propose to clip IK solutions that pass through the desired end-effector position to be within pre-specified limits, doing so alleviates the risk of getting bad but feasible IK solutions.
When a ball is launched, the desired hit point for conditioning is identified by computing where the predicted ball trajectory crosses a pre-specified hit plane, and this point is then transformed into the frame of the mobile base. Iterative IK updates to the available Degrees of Freedom~(DoFs) are performed to reach this desired hit point, and the total updates are clipped to be within the specified limits. The pseudocode of this procedure is presented in \Cref{app:stroke-controller}.

Second, based on the observation from \cite{zaidi2022athletic} that early positioning movement for tennis-playing robots can improve the chances of a successful return, we allow the wheelchair to continuously adjust its position during the conditioning process and execute the stroke independently based on the anticipated ball arrival time. A flow chart explaining the state-machine of the stroke controller is illustrated in \Cref{app:stroke-controller}.



\subsection{Refining primitives through human feedback}
\label{subsec:refine-prim}
In~\cite{gomez2020adaptation}, the parameters of ProMP are trained through a dataset of successful demonstrations obtained through kinesthetic teaching or engineered controllers. While the ProMP model is initially trained to recreate a demonstrated behavior (i.e., a robot arm trajectory that hits a ball), the result is suboptimal due to both the suboptimality of the demonstration itself, out-of-distribution incoming flights of the tennis ball and the hardware-software system's inability to perfectly execute a commanded trajectory. 

Nonetheless, the trained primitive does serve as an excellent starting point for exploring better trajectories that can be executed on the hardware. Therefore, we propose to iteratively improve the primitives by having a human evaluator assign scalar feedback indicative of the quality of the executed trajectory. We collect human feedbacks $\{r_n\}_{n=1}^N$ from the evaluator for $N$ executed trajectories $\{\{ q_{nt} \}_{t=1}^{T_n}\}_{n=1}^N$ and then construct a dataset $\mathcal{D}$ which consists of trajectories and the associated importance weights ($\alpha_n$, obtained as softmax over feedbacks) to optimize the following weighted log-likelihood objective: 
\vspace{-25pt}

\begin{align}
\vspace{-45pt}
    \texttt{WeightedLogLikelihood}(\mathcal{D} \mid \theta) = \sum_{i=1}^N \alpha_n \log(\texttt{Likelihood}(\;\{q_{nt}\}_{t=1}^{T_n} \mid \theta\;))
\end{align}

The parameters are optimized with the EM procedure outlined in \cite{gomez2020adaptation}. In the M-step we compute the weighted average of the estimates from the E-step as a consequence of the weighted log-likelihood objective. This algorithm can be considered a part of the TAMER\cite{knox2009interactively} class of algorithms, as we are interactively shaping the distribution of executed trajectories using human feedback.


\begin{algorithm}
        \caption{Iterative refinement of ProMP parameters}
        \label{alg:iter-emwll}
        \setstretch{1.15}
        \begin{algorithmic}
        \Require ProMP parameters $\theta$ = ($\mu_w$, $\Sigma_w$, $\Sigma_y$) trained from human demonstrations 
        \Repeat
            \State Execute N trajectories $\{\tau_n\}$ from conditioned execution of $\theta$ and obtain human feedback $\{r_n\}$
            \State Compute importance weights over trajectories, $\{\alpha_n\} \gets \texttt{SoftMax}_{n=1}^N(\{r_n\})$ 
            \State With dataset $\mathcal{D} = \{ (\tau_n, \alpha_n) \}$, perform $\theta \gets \texttt{EM-WeightedLogLikelihood}(\mathcal{D}, \theta_{\text{init}} = \theta)$ 
        \Until {convergence}
        \end{algorithmic}
        \label{alg:refine}
\end{algorithm}

\subsection{Experiment Setup}
\label{sec:exp-setup}
\begin{figure}
    \centering
    \label{fig:collection-overview}
    \begin{subfigure}[b]{0.4\textwidth}
        \centering
        \includegraphics[width=\textwidth]{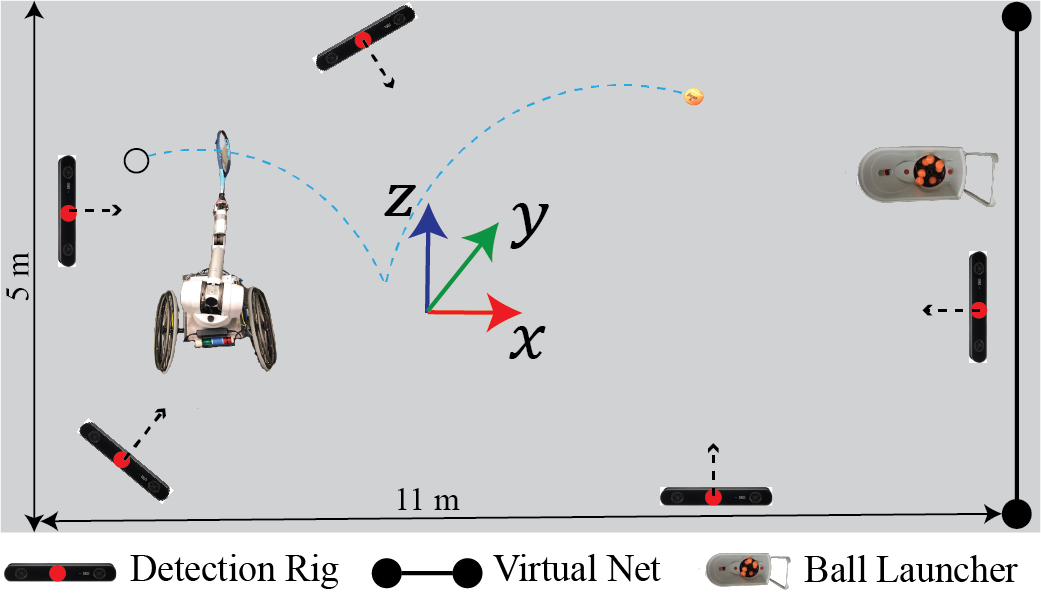}
        \caption{}
        \label{subfig:lab-setup}
    \end{subfigure}\hspace{5mm}
    \begin{subfigure}[b]{0.4\textwidth}
        \centering
        \includegraphics[height=3.2cm]{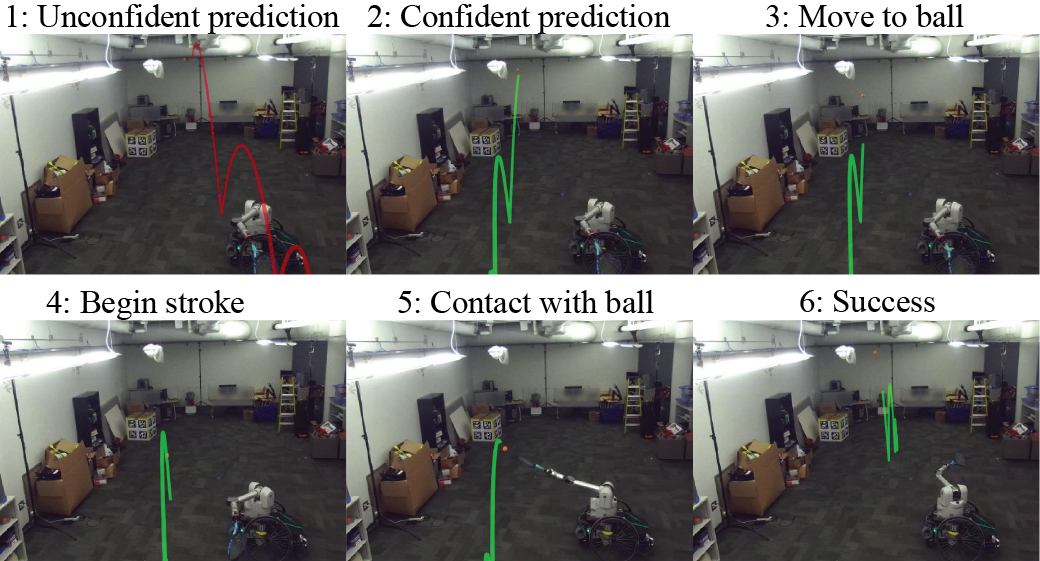}
        \caption{}
        \label{subfig:example-hit}
    \end{subfigure} \hfill 
    \caption{(a) Overview of the lab setup. (b) Wheelchair Tennis Robot executing a stroke.}
\end{figure}

For our experiments, we evaluate the performance of the ProMP stroke controller in the lab setup illustrated in \Cref{subfig:lab-setup}. We initialize the ProMP parameters by training it on a dataset of successful demonstration obtained from a manually engineered stroke (\Cref{subfig:example-hit}); this serves as our base primitive. To evaluate the proposed fine-tuning algorithm (\Cref{subsec:refine-prim}) we collect a dataset by running the base primitive. For each ball launched, we record the joint states to a ROS\cite{quigley2009ros} bag file and store the associated feedback based on criteria listed in \Cref{tab:reward-criteria}. We segment the trajectories from the bag files by analyzing the maximum of all joint velocities to determine points of inflection, this constitutes the start and end of the trajectory. The hit phase parameter is chosen based on where the end-effector approximately crosses the pre-specified hit plane. \Cref{fig:traj-seg} illustrates the process of trajectory extraction. We also evaluate the impact of the number of trajectories used for refining the primitives by comparing the performance of the primitives trained on datasets of 20 and 50 trajectories. Performance measures hit rate, success rate (success corresponds to good hits), and the average reward on a consecutive sequence of ball launches.

\begin{table}
    \begin{minipage}{0.5\linewidth}
		\centering
		\includegraphics[height=3cm]{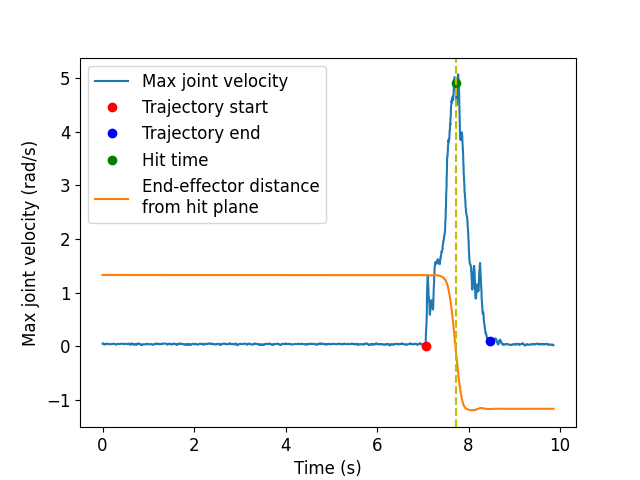}
        \captionof{figure}{Example of trajectory segmentation from recorded joint states.}
        \label{fig:traj-seg}
	\end{minipage}
	\begin{minipage}{0.5\linewidth}
		\label{tab:reward-criteria}
		\centering
		\begin{tabular}{p{3.5cm}|l}
        \hline
         {\bf criteria} & {\bf reward} \\[0.2em] \hline
        miss by a large margin &  0 \\[0.2em] 
        miss but close ($\leq$ 5\cm) & 0.25 \\[0.2em] 
        hit but not good enough & 0.5 \\[0.2em] 
        good hit (hit side pillar) & 1 \\[0.2em] 
        good hit (above the net) & 2 \\ \hline
    \end{tabular}
    		\vspace{1em}\caption{Criteria of human feedback}
	\end{minipage}\hfill
\end{table}

\section{Results}
\label{sec:results}
We report the base primitive performance and the performance post-refinement in \Cref{tab:result} based on the setup described in \Cref{sec:exp-setup}. Performance is reported over a consecutive run of 10 balls. 
\begin{table}[!ht]
    \centering
    \begin{tabular}{l|c|cc}
        \hline
    {\bf \# trajectories} & 0 (base primitive) & 20 & 50 \\[0.2em] \hline
        {\bf Hit Rate} &  60\% &  40\%  & 50\%   \\[0.2cm] \hline
        {\bf Success Rate} & 40\% & 40\%   & 40\%   \\[0.2cm] \hline
        {\bf Avg. Reward} & 0.75 &   0.85   & 0.85   \\[0.2cm] \hline
    \end{tabular}
    \vspace{0.3cm}
    \caption{Performance over the number of trajectories used for refining the base primitive.}
    \label{tab:result}
\end{table}
\vspace{-10mm}
\section{Conclusion}
\label{sec:conclusion}
We have successfully demonstrated a safely executed ground strike primitive on a wheelchair tennis robot. 
We proposed a formulation to fine-tune learned primitives online and conducted an evaluation. While we do not observe significant improvements in success rates with fine-tuning, we do see a small increase in the average feedback rewarded to the primitives, so most balls have been missed by small margins. Future work in this direction can consider different reward designs and study how the choice of the reward impacts the learned primitive.

\textbf{Future Work}
In the future, we would like to explore methods to improve upon learned primitives to 1) ensure safe behavior and 2) increase task performance. Improving learned robot behavior has relations across Active Learning with Human Feedback \cite{ARGALL2009469}, Reinforcement Learning of robot skills \cite{Ibarz2021HowTT, Carvalho2022ResidualRL}, and Human-Robot Interaction. We would additionally like to consider several feedback paradigms in improving robot motion including natural language, kinesthetic teaching, third-person demonstration, etc.

\clearpage

\bibliography{references}  

\newpage
\appendix

\section{Kinematic Model of the Robot}
\label{app:kinematic-model}
\begin{figure}[H]
    \caption{The kinematic model used for hit-point conditioning of primitive}
    \centering
    \includegraphics[width=0.55\textwidth]{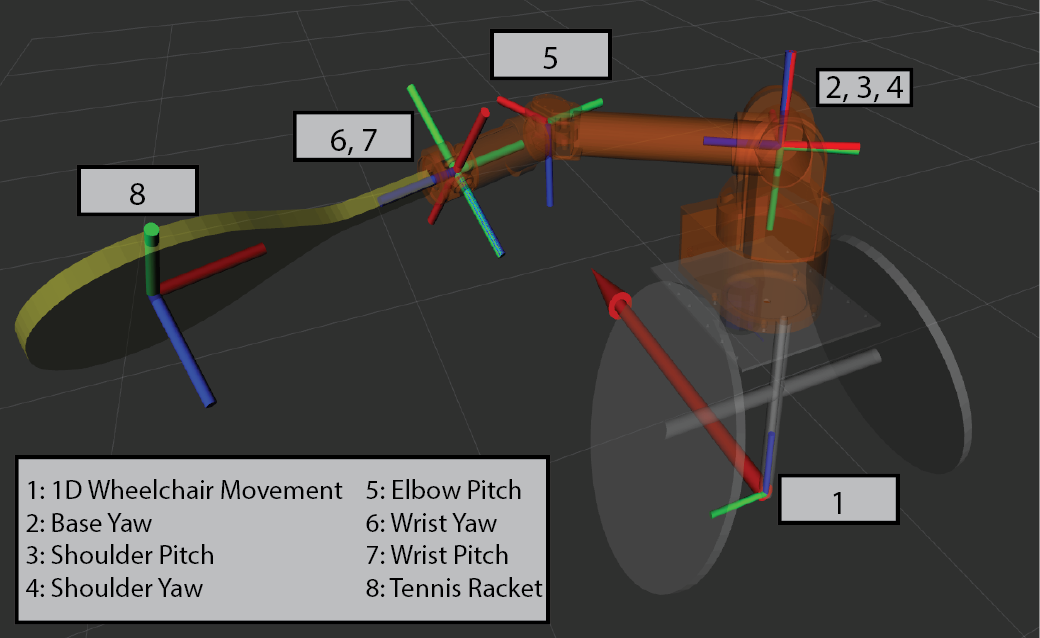}
    \label{fig:kinematics-model}
\end{figure}

\section{Details of the stroke controller}
\label{app:stroke-controller}

\begin{figure}[H]
    \centering
    \caption{Flowchart depicting the state-machine of the stroke controller}
    \includegraphics[width=0.97\textwidth]{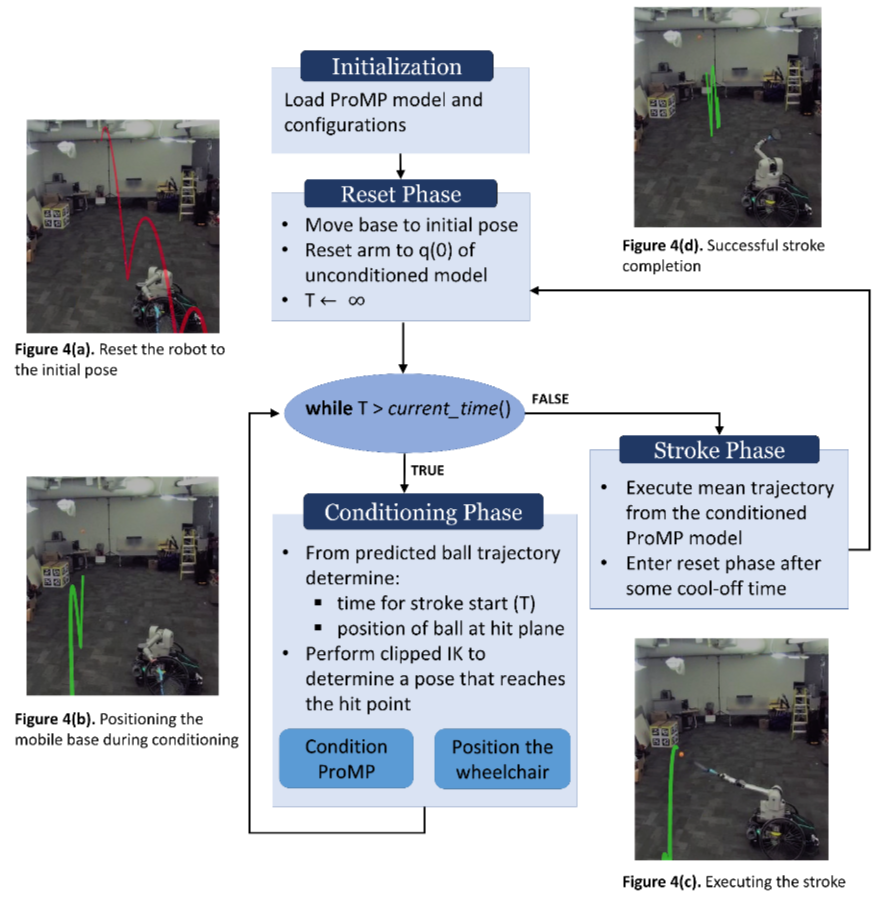}
    \label{fig:stroke-controller-flowchart}
\end{figure}

\begin{algorithm}[H]
        \caption{Clipped Inverse-Kinematics for conditioning of stroke primitives}
        \label{algo:cik}
        \setstretch{1.05}
        \begin{algorithmic}[1]
        \Require 
            \Statex $\bullet\,$ Desired hit point position $x_d$ in local mobile-base frame 
            \Statex $\bullet\,$ seed configuration $q_{\text{seed}}$
            \Statex $\bullet\,$ Forward kinematics model $f: (r, q) \to \R^3$ and the corresponding manipulator jacobian $J(r,q)$, where $r$ is the wheelchair position and $q$ is the joint state 
            \Statex $\bullet\,$ lower (\texttt{LL}) and upper (\texttt{UL}) limits of allowed wheelchair and joint movements. 
        \Statex
        \State Let $q_c = q_{\text{seed}}$ 
        \State Current end-effector position $x_c = f(q_c, 0)$  
        \State Let $\texttt{net\_dq} = 0$ \Comment{\parbox[t]{2.7in}{\raggedright required joint position offsets from the seed state}}
        \State Let $\texttt{net\_dr} = 0$ \Comment{\parbox[t]{2.7in}{\raggedright net movement required for wheelchair}}
        \Repeat
            \State $\Delta r, \Delta q \gets J(\texttt{net\_dr}, q_c)^\dagger (x_d - x_c)$
            \State $\texttt{net\_dr}\,, \texttt{net\_dq} \gets \texttt{ClipedIncrement}(\texttt{net\_dr}\,, \texttt{net\_dq}, \Delta r, \Delta q, \texttt{LL}, \texttt{UL})$
            \State $q_c \gets q_{\text{seed}} + \texttt{net\_dq}$
            \State $x_c \gets f(\texttt{net\_dr}, q_c)$
        \Until iteration $<$ N {\bf and} !\texttt{allclose}($x_c$, $x_d$, $\epsilon$)
        \Statex
        \State {\bf command} wheelchair to move: \texttt{net\_dr} 
        \State {\bf condition} stroke primitive to pass through: $q_{\text{seed}} + \texttt{net\_dq}$
        \end{algorithmic}
\end{algorithm}

\end{document}